%% file: ECML_714.tex
\documentclass[runningheads]{llncs}
\usepackage{graphicx}
\usepackage{algorithm}
\usepackage{algorithmic}

\usepackage{newfloat}
\usepackage{listings}

\usepackage{textcase}
\usepackage{dirtytalk}
\usepackage{etoolbox}
\usepackage{times}
\usepackage{soul}
\usepackage{url}
\usepackage[hidelinks]{hyperref}
\usepackage[utf8]{inputenc}
\usepackage[small]{caption}
\usepackage{graphicx}
\usepackage{amsmath}

\usepackage{amssymb}
\usepackage{bm}
\usepackage{booktabs}
\usepackage{algorithm}
\usepackage{algorithmic}
\usepackage{xcolor}
\usepackage{yhmath}
\usepackage{bm}
\usepackage{bbm}
\usepackage{mymacros}

\newcommand{\argmax}{\arg\!\max}

\usepackage{thmtools,thm-restate}
\def\I{\mathrm{I}}

\usepackage{thmtools}
\usepackage{thm-restate}
\usepackage{datetime}
\begin{document}
\title{On Projectivity in Markov Logic Networks}

%
\author{Sagar Malhotra \inst{1,2} \and
Luciano Serafini\inst{1}}
\authorrunning{S. Malhotra  et al.}
%
\institute{Fondazione Bruno Kessler, Italy \and
University of Trento, Italy}
\maketitle              
\input{00_abstract}
\input{01_introduction}
\input{02_related}
\input{03_background}
\input{04_MLN}
\input{05_projectivity}

\input{06_rbm}
\input{07_comparison}

\input{08_learning}
\input{09_conclusion}


\bibliographystyle{splncs04}
\bibliography{ECML_714}
\input{appendix}

\end{document}

%% file: 00_abstract.tex
\begin{abstract}
Markov Logic Networks (MLNs) define a probability distribution on relational 
structures over varying domain sizes. Like most relational models, MLNs do not admit consistent marginal inference over varying domain sizes. Furthermore, MLNs learned on a fixed domain do not generalize to domains of varied sizes. In recent works, connections have emerged between  domain size dependence, lifted inference, and learning from  a sub-sampled domain. The central idea of these works is the notion of  \emph{projectivity}. The probability distributions ascribed by projective models render the marginal probabilities of sub-structures independent of the domain cardinality. Hence, projective models admit efficient marginal inference. Furthermore, projective models potentially allow efficient and consistent parameter learning from sub-sampled domains. In this paper, we characterize the necessary and sufficient conditions for a two-variable MLN to be projective. We then isolate a special model in this class of MLNs, namely Relational Block Model (RBM). We show that, in terms of data likelihood maximization, RBM is the best possible projective MLN in the two-variable fragment. Finally, we show that RBMs also admit consistent parameter learning over sub-sampled domains.

\end{abstract}


%% file: 01_introduction.tex
\section{Introduction}
Statistical Relational Learning \cite{SRL_LISA,SRL_LUC} (SRL) is concerned with representing and learning probabilistic models over relational structures. In  recent years, the behavior of SRL models under varying domain sizes has come into significant attention \cite{Projectivity_first,Population_Extrapolation}. Many works have observed that SRL frameworks exhibit unwanted behaviors over varying domain sizes \cite{Population_Extrapolation,DA_MLN}.  These behaviors make models learned from a sub-sampled domain unreliable for inference over larger (or smaller) domain sizes \cite{DA_MLN}. Drawing on the works of Shalizi and Rinaldo \cite{Projectivity_Rinaldo} on Exponential Random Graphs (ERGMs), Jaeger and Schulte \cite{Projectivity_first} have recently introduced the notion of  \emph{projectivity} as a strong form of guarantee for good scaling behavior in SRL models. A projective model requires that the probability of any given query, over arbitrary $m$ domain objects, is completely independent of the domain size. Jaeger and Schulte \cite{Projectivity_first}  identify restrictive fragments of many SRL models to be projective. But wether these fragments are complete characterization of projectivity, remains an open problem.

In this paper our goal is to characterize projectivity for a specific class of SRL models, namely Markov Logic Networks (MLNs)\cite{richardson2006markov}. MLNs are amongst the most prominent template based SRL models. An MLN is a Markov Random Field with features defined in  terms of function-free weighted First Order Logic (FOL) formulae. Jaeger and Schutle \cite{Projectivity_SRL} show that an MLN is projective if:  any pair of atoms in each of  its FOL formulae share the same set of variables. In this paper, we show that this characterization is not complete. Furthermore, we completely characterize projectivity for the class of MLNs with at-most 2 variables in their formulae. A key contribution of our work is that we are able to provide a parametric restriction which can be easily incorporated into any MLN learning algorithm. Furthermore, we are able to identify a special class of projective MLNs, namely Relational Block Models (RBMs), which maximally maximize the training data likelihood w.r.t any other projective MLN in the two variable fragment. Finally, we show that RBMs admit consistent maximum likelihood parameter estimation over varying domain sizes.

The paper is organized as follows: We first contextualize our work w.r.t the related works in this domain. We then provide some background and notation on FOL and relational structures. We also elaborate on the fragment of FOL with at most two variables i.e. FO$^2$ and define the notion of  FO$^2$ interpretations as multi-relational graphs. We also overview some results on Weighted First Order Model Counting. In the subsequent section, we provide a  parametric representation for any MLN in the two variable fragment. We then dedicate a section to the main result of this paper i.e. the necessary and sufficient conditions for an MLN in the two variable fragment to be projective. Based on the projectivity criterions we identify a special class of models namely Relational Block Models. We dedicate a complete section to RBMs and elaborate on their useful properties. Finally, we discuss the consistency and efficiency aspects of learning for projective MLNs and RBMs.


%% file: 02_related.tex
\section{Related work}
Projectivity has emerged as a formal notion of interest through multiple independent lines of works across ERGM and SRL literature. The key focus of these works have been analyzing \cite{Proj_TOM,Population_Extrapolation} or mitigating \cite{A_MLN,DA_MLN} the
effects of varying domain sizes on relational models. The major step in formalizing the notion of projectivity  can be attributed to Shalizi and Rinaldo \cite{Projectivity_Rinaldo}. The authors both formalize and characterize the sufficient and necessary conditions for ERGMs to be projective. It is interesting to 
note that their projectivity criterion is strictly structural i.e. they put no restrictions on parameter values but rather inhibit the class of features that can be defined as sufficient statistics in ERGMs. In contrast our results w.r.t MLN are strictly parametric (which may at times correspond to structural restrictions as well). With respect to SRL, the notion of projectivity was first formalized by Jaeger and Schulte \cite{Projectivity_first}, they provide  many restrictive fragments of SRL models to be projective. In \cite{Projectivity_SRL}, Jaeger and Schulte, significantly extend the scope of projective models by characterizing necessary and sufficient conditions for an arbitrary model on relational structures to be projective. Their characterization is expressed in terms of the so called AHK models. But as they conclude in \cite{Projectivity_SRL}, expressing AHK models in existing SRL frameworks remains a significant open challenge. Hence, a complete characterization of projectivity in most SRL languages is still an open challenge. In \cite{Felix_Weit}, Weitkamper has  shown that the characterization of projectivity provided by Jaeger and Schulte \cite{Projectivity_first}, for probabilistic logic programs under distribution semantics, is indeed complete. In this work, we will extend this characterization to two variable fragment of Markov Logic Networks.

Another correlated problem  to projectivity is learning from sub-sampled or smaller domains. In the relational setting projectivity is not a sufficient condition for 
consistent learning from sub sampled domains \cite{Projectivity_first}. Mittal et. al.  have proposed a solution to this problem by introducing domain-size dependent scaling down factors \cite{DA_MLN} for MLN weights. Although empirically effective, the scaling down factors are not  shown to be a statistically sound solution. On the other hand, Kuzelka et. al. \cite{Relational_Marginal_Polytope}, provide a statistically sound approach to approximately obtain the correct distribution for a larger domain but there approach 

%% file: 03_background.tex
\section{Background}
\subsubsection{Basic Definitions.}We use the following basic notation. The set of integers $\{1,...,n\}$ is denoted by $[n]$. 
We use $[m:n]$ to denote the set of integers $\{m,...,n\}$. Wherever the set of integers $[n]$ is obvious from the context we will use $[\overline{m}]$ to represent the set $[m+1:n]$. We use $\bm{k} = \langle k_1,...,k_m \rangle$ to denote an $n$-partition i.e. $k_i \in \mathbb{Z^{+}}$ and  $\sum_{i\in[m]}k_i = n$.
We will also use multinomial coefficients denoted by  $$\binom{n}{k_1,...,k_m} =  \binom{n}{\bk}= \frac{n!}{\prod_{i\in [m]} k_i !}$$

\subsubsection{First Order Logic and Relational Substructures.}We assume a function-free First Order Logic (FOL) language
$\mathcal{L}$ defined by a set of variables $\mathcal{V}$ and a set of
relational symbols $\mathcal{R}$. We use $\Delta$ to denote a domain of
$n$ constants. For $a_1,...,a_k \in \mathcal{V} \cup \Delta$ and $R\in \mathcal{R}$, we call $R(a_1,...a_k)$ an \emph{atom}. A \emph{literal} is an atom or the negation of an atom. If $a_1,...,a_k \in \mathcal{V}$, then the atom is called a \emph{first order atom}, whereas if $a_1,...,a_k \in \Delta$, then it's called a \emph{ground atom}. We use $\mathcal{F}$ to denote the set of first order atoms and $\mathcal{G}$ to denote the set of ground atoms. A \emph{world} or an \emph{interpretation} $\omega:\mathcal{G} \rightarrow \{\mathbf{T},\mathbf{F}\}$ is a function that maps each ground atom to a boolean.  The set of interpretations $\omega$ in the language $\mathcal{L}$ and the domain $\Delta$ of size $n$ is denoted by $\Omega^{(n)}$. We say that $\omega \in \Omega^{(n)}$, has a size $n$ and is also called an $n$-world. For a subset $\mathrm{I}\subset \Delta$, we use $\omega \downarrow \mathrm{I}$ to denote the partial interpretation induced by $\mathrm{I}$. Hence, $\omega \downarrow \mathrm{I}$  is an interpretation over the ground atoms containing only the domain elements in $\mathrm{I}$. 
\begin{example}
    Let us have a language with only one relational symbol $R$ of arity $2$  and a domain $\Delta = \{a,b,c\}$. Let us have the following interpretation $\omega$:
    \begin{center}
        \begin{tabular}{ |c|c|c|c|c|c|c|c|c| } 
         \hline
         $R(a,a)$ & $R(a,b)$ & $R(a,c)$ &  $R(b,a)$ & $R(b,b)$ & $R(b,c)$  &  $R(c,a)$ & $R(c,b)$ & $R(c,c)$\\
         \hline 
         $\mathbf{T}$ & $\mathbf{T}$ & $\mathbf{F}$ & $\mathbf{T}$ & $\mathbf{T}$ & $\mathbf{F}$ & $\mathbf{T}$ & $\mathbf{T}$ & $\mathbf{F}$ \\ 
         \hline
        \end{tabular}
    \end{center}
then $\omega \downarrow \{a,b\}$ is given as:

\begin{center}
    \begin{tabular}{ |c|c|c|c| } 
     \hline
     $R(a,a)$ & $R(a,b)$ & $R(b,a)$ &  $R(b,b)$\\
     \hline 
     $\mathbf{T}$ & $\mathbf{T}$ & $\mathbf{F}$ & $\mathbf{T}$\\ 
     \hline
    \end{tabular}
    \end{center}

\end{example}
For most of our purposes, we will be able to assume w.l.o.g that $\Delta = [n]$.

\subsubsection{FO$^2$, $m$-Types and $m$-Tables.}FO$^2$ is the fragment of FOL with two variables. We will use the notion of $1$-types, $2$-type, and $2$-tables as presented in \cite{kuusisto2018weighted}. A $1$-type is a conjunction of a maximally consistent set of first order literals containing only one variable. For example, in an FO$^2$ language on the unary predicate $A$ and binary predicate $R$,  ${A(x)\land R(x,x)}$ and $ {A(x)\land \neg R(x,x)}$ are examples of 1-types in variable $x$. A 2-table is a conjunction of maximally consistent first order literals containing exactly two distinct variables. Extending the previous example, ${R(x,y)\land \neg R(y,x)}$ and ${R(x,y)\land R(y,x)}$ are instances of 2-tables.
We assume an arbitrary order on the 1-types and 2-tables, hence, we use ${i(x)}$ to denote the $i^{th}$ 1-type and $ {l(x,y)}$ to denote the $l^{th}$ 2-table. Finally, a $2$-type is a conjunction of the form 
${i(x)\land j(y) \land l(x,y) \land (x \neq y)}$ and we use $ {ijl(x,y)}$ to represent it. In a given interpretation $\omega$, we say a constant $c$ realizes the $i^{th}$ 1-type if $ {\omega \models i(c)}$, we say a pair of constants $(c,d)$ realizes the $l^{th}$ 2-table if $ {\omega \models l(c,d)}$ and $(c,d)$ realizes the 2-type $ {ijl(x,y)}$ if $ {\omega \models ijl(c,d)}$. We call the 2-type ${ijl(y,x)}$ the dual of ${ijl(x,y)}$ and denote it by $ {\underline{ijl}(x,y)}$. We will use $u$ to denote the number of 1-types and $b$ to denote the number of 2-tables in a given FO$^2$ language.

\subsubsection{Interpretations as Multi-relational Graphs.}Given an
FO$^2$ language $\mathcal{L}$ with interpretations defined over the
domain $\Delta = [n]$, we can represent an interpretation
$\omega \in \Omega^{(n)}$ as a multi-relational graph $(\x,\y)$. This
is achieved by defining $\x = (x_1,...,x_n)$ such that $x_q=i$ if
$ {\omega \models i(q)}$ and by defining
$\y = (y_{12},y_{13},...y_{qr},...,y_{n-1,n})$, where $q<r$, such that
$y_{qr} = l$ if $ {\omega \models l(q,r)}$. We also define
$k_i = k_i(\x) = k_i(\omega) := |\{c \in \Delta: c \models i(c) \}| $, $h^{ij}_l = h^{ij}_l(\y) = h^{ij}_l(\omega) := |\{(c,d) \in
  \Delta^{2}: \omega \models ijl(c,d)\} |$ and
  for any $D\subseteq\Delta^2$, $h^{ij}_{l}(D) = h^{ij}_{l}(\omega,D):=|\{ (c,d):\omega \models ijl(c,d) \text{ and } (c,d)
  \in D\}|$. Notice that
  $\sum_{i\leq j}\sum_{l\in[b]} h^{ij}_l = \binom{n}{2}$ and
  $\sum_{l\in [b]}h^{ij}_{l} = \bk(i,j)$, where $\bk(i,j)$ is defined
  in equation \eqref{kij} . We use
$(\x_{\mathrm{I}} , \y_{\mathrm{I}})$ to represent the
multi-relational graph for $\omega \downarrow \mathrm{I}$. Throughout
this paper we will use an interpretation $\omega$ and it's
multi-relational graph $(\x,\y)$ interchangeably.

\subsubsection{Weighted First Order Model Counting in FO$^2$.}We will briefly review Weighted First Order Model Counting (WFOMC) in FO$^2$ as presented in \cite{FOMC_arxiv}. WFOMC is formally defined as follows:
\begin{equation*}
    \wfomc(\Phi,n) := \sum_{\omega \in \Omega^{(n)}:\omega \models \Phi}w(\omega)
\end{equation*}
where $\Phi$ is an FOL formula, $n$ is the size of the domain and $w$ is a weight function that maps each interpretation $\omega$ to a positive real. First Order Model Counting (FOMC) is the special case of WFOMC, where for all $\omega \in \Omega^{(n)}$, $w(\omega) = 1$. We assume that 
$w$ does not depend on individual domain constants, which implies that $w$ assigns same weight to two interpretations which are isomorphic under the permutation of domain elements. 

A universally quantified FO$^2$ formula $\forall xy. \Phi(x,y)$ can be equivalently expressed as $\forall xy. \Phi(\{x,y\})$, where $\Phi(\{x,y\})$ is defined as $\Phi(x,x)\land \Phi(x,y)\land \Phi(y,x)\land \Phi(y,y)\land( x \neq y)$. A \emph{lifted interpretation} denoted by $\tau: \mathcal{F} \rightarrow \{\mathbf{T,F}\}$ assigns boolean values to first order atoms. The truth value of the quantifier free formula $\Phi(x,y)$ under a lifted interpretation $\tau$, denoted by $\tau(\Phi(x,y))$, is computed by applying classical semantics of the propositional connectives to the truth assignments of atoms of $\Phi(\{x,y\})$ under $\tau$. We then define 
\begin{equation}
    \label{eq: n_ijl}
    n_{ijl} := \left|\left\{\tau\mid\tau\models\Phi(\{x,y\})\wedge ijl(x,y) \right\}\right|
\end{equation}
and $n_{ij} := \sum_{l\in [b]} n_{ijl}$. First Order Model Counting for a universally quantified formula $\forall xy. \Phi(x,y)$ is then given as:
\begin{align}
    \label{eq:fomc-universal-form}
    \fomc(\forall xy. \Phi(x,y),n) = &
   \sum_{\bk}  \binom{n}{\bm{k}}
   \prod_{\substack{{i\leq j}\\{i,j\in[b]}}}\!\!\! n_{i j}^{\bk(i,j)}
\end{align}
where $\bk =\left<k_1,\dots,k_u\right>$ is a $u$-tuple of non-negative integers, $\binom{n}{\bm{k}}$ is the multinomial coefficient and
\begin{align}
    \label{kij}
  \bk(i,j) =
  \begin{cases}
      \frac{k_{i}(k_{i} - 1)}{2} & \text{if $i=j$} \\
       k_{i}k_{j} & \text{otherwise} \\
     \end{cases}
\end{align}
Intuitively, $k_i$ represents the number of constants $c$ of 1-type $i$. Also a given constant realizes exactly one 1-type. Hence, for a given 
$\bk$, we have $\binom{n}{\bk}$ possible ways of realizing $k_i$ 1-types. Furthermore, given a pair of constants $c$ and $d$ such that 
   $c$ is of 1-type $i$ and $d$ is of 1-type $j$, the number of extensions to the binary predicates containing both $c$ and $d$, such that the extensions are a model of $\forall xy. \Phi(x,y)$, is given by $n_{ij}$ independently of all other constants. Finally, the exponent $\bk(i,j)$ accounts for all possible pair-wise choices of constants given a $\bk$ vector.  
Equation \eqref{eq:fomc-universal-form} was originally proven in \cite{Symmetric_Weighted}, we refer the reader to \cite{FOMC_arxiv} for the formulation presented here.
  
\subsubsection{Families of Probability Distributions and Projectivity.} We will be interested in probability distributions over the set of interpretations or equivalently their multi-relational graphs.
A family of probability distributions $\{P^{(n)}: n \in \mathbb{N}  \}$ specifies, for each finite domain of size $n$, a distribution $P^{(n)}$ on the possible $n$-world set $\Omega^{(n)}$ \cite{Projectivity_SRL}. We will mostly work with the so-called exchangeable probability distributions \cite{Projectivity_SRL} i.e. distributions where $P^{(n)}(\omega) = P^{(n)}(\omega') $ if $\omega$ and $\omega'$ are isomorphic. A distribution $P^{(n)}(\omega)$ over $n$-worlds induces a marginal probability distribution over  $m$-worlds $\omega' \in \Omega^{(m)}$ as follows:
\begin{equation*}
    P^{(n)}\downarrow [m](\omega') = \sum_{\omega \in \Omega^{(n)}:\omega \downarrow [m] =\omega'} P^{(n)}(\omega)
\end{equation*}
Notice that due to exchangeability $P^{(n)} \downarrow \mathrm{I}$ is the same for all subsets $\I$ of size $m$, hence we can always assume any induced $m$-world  to be $\omega \downarrow [m]$. We are now able to define projectivity as follows:
\begin{definition}[\cite{Projectivity_SRL}]
    An exchangeable family of probability distributions is called projective if for all $m < n$: $$P^{(n)} \downarrow [m] = P^{(m)}$$
\end{definition}
when dealing with probability distributions over multi-relational representation, we denote by $(\bX,\bY)$ the
random vector where, $\bX=(X_1,\dots,X_n)$ and each $X_i$ takes value in $[u]$;
and $\bY=(Y_{12},Y_{13},\dots,Y_{qr},\dots,Y_{n-1,n})$ where $q < r$ and $Y_{qr}$ takes values in $[b]$.


%% file: 04_MLN.tex
\section{Markov Logic Networks: A Parametric Normal Form}
A Markov Logic Network (MLN) $\Phi$ is defined by a set of  weighted
formulas $\{(\phi_i,a_i)\}_i$, where $\phi_i$ are quantifier free, function-free
FOL formulas with weights ${a_i \in \mathbb{R}}$. An MLN $\Phi$ induces a
probability distribution over the set of possible worlds $\omega \in \Omega^{(n)}$: 
\begin{equation*}
    P^{(n)}_{\Phi}(\omega) = \frac{1}{Z(n)}\exp\Bigl(\sum_{(\phi_i,a_i)\in \Phi}a_i.N(\phi_i,\omega)\Bigr)
\end{equation*}
where $N(\phi_i,\omega)$ represents the number of true groundings of
$\phi_i$ in $\omega$. The normalization constant $Z(n)$ is called the
\emph{partition function} that ensures that $P^{(n)}_{\Phi}$ is a
probability distribution.


\begin{theorem}
    \label{th: MLN_Normal_Form}
    Any Markov Logic Network (MLN) $\Phi = \{(\phi_i,a_i)\}_i$ on a domain of size $n$, such that $\phi_i$ contains at-most two variables, can be expressed as follows:
\begin{equation}
    \label{eq:mln_normal_form}
P^{(n)}_{\Phi}(\omega) = \frac{1}{Z(n)} \prod_{i \in [u]} s_{i}^{k_i} \prod_{\substack{i,j\in [u]\\ i \leq j}}\prod_{l \in [b]} (t_{ijl})^{h^{ij}_l}
\end{equation}
where $s_i$ and $t_{ijl}$ are positive real numbers and $k_i$ is $k_i(\omega)$ and $h^{ij}_l$ is equal to $h^{ij}_l(\omega)$. 
\end{theorem}

\begin{proof}
  Let $\Phi = \{(\phi_i,a_i)\}_i$  be an MLN, such that $\phi_i$ contains at-most two variables.
  Firstly, every weighted formula  $(\phi(x,y),a)\in\Phi$ that contains exactly two variables
  is replaced by two weighted formulas $(\phi(x,x),a)$ and $(\phi(x,y)\land (x \neq y) ,a)$.
  The MLN distribution $P_{\Phi}^{(n)}$ is invariant under this transformation. Hence, $\Phi$ can be equivalently written as $\{(\alpha_q(x),a_q)\}_{q} \cup \{(\beta_p(x,y),a_p)\}_{p} $, where $\{\alpha_q(x)\}_{q}$ is the set of formulas containing only the variable $x$ and $\{\beta_p(x,y)\}_{p}$ is the set of formulas containing both the variables $x$ and $y$. Notice that every $\beta_p(x,y)$ entails $x\neq y$.

Let us have  $\omega \in \Omega^{(n)}$, where we have a domain constant $c$ such that $\omega\models i(c)$. Now notice that the truth value of ground formulas $\{\alpha_q(c)\}_q$ in $\omega$ is completely determined by $i(c)$ irrespective of all other domain constants. Hence, the (multiplicative) weight contribution of $i(c)$ to the weight of $\omega$ can be given as $\exp(\sum_{q} a_q\mathbbm{1}_{i(x) \models \alpha_q(x)})$. We define $s_i$ as follows:
\begin{equation}
    \label{si}
    s_i  = \exp\bigl(\sum_{q}a_q\mathbbm{1}_{i(x) \models \alpha_q(x)}\bigr)
\end{equation} 
Clearly, this argument can be repeated for all the domain constants realizing any 1-type in $[u]$. Hence, the (multiplicative) weight contribution due to 1-types of all domain constants and equivalently due to the groundings of all unary formulas, is given as $\prod_{i\in[u]}s_i^{k_i}$. 

We are now left with weight contributions due to the binary formulas $\{(\beta_p(x,y),a_p)\}_{p}$. Due to the aforementioned transformation, each binary formula $\beta(x,y)$ contains a conjunct $(x \neq y)$. Hence, all groundings of $\beta(x,y)$ such that both $x$ and $y$ are mapped to the same domain constants evaluate to false. Hence, we can assume that $x$ and $y$ are always mapped to distinct domain constants. Let us have an unordered pair of domain constants $\{c,d\}$ such that $\omega\models ijl(c,d)$. The truth value of any binary ground formula $\beta(c,d)$ and $\beta(d,c)$ is completely determined by $ijl(c,d)$ irrespective of all other domain constants. Hence, the multiplicative weight contribution due to the ground formulas $\{\beta_p(c,d)\}_p \cup \{\beta_p(d,c)\}_p$ is given as $t_{ijl}$, where $t_{ijl}$ is defined as follows:
\begin{equation}
    \label{t_ijl}
t_{ijl} =\exp\left(\sum_{p}a_p\mathbbm{1}_{ijl(x,y) \models \beta_p(x,y)} + \sum_{p}a_p\mathbbm{1}_{ijl(x,y) \models \beta_p(y,x)}\right)
\end{equation}


Hence, the weight of an interpretation $\omega$ under the MLN $\Phi$ is given as  
$$\prod_{i \in [u]} s_{i}^{k_i} \prod_{\substack{i,j\in [u]\\ i \leq j}}\prod_{l \in [b]} (t_{ijl})^{h^{ij}_l}$$
\end{proof}
\begin{definition}
    Given a MLN in the parametric normal form given by equation \eqref{eq:mln_normal_form}. Then $f_{ij}$ is defined as follows:
    \begin{equation}
        f_{ij} = \sum_{l\in [b]}t_{ijl}
    \end{equation}
\end{definition}

We will now provide the parameterized version of the partition function $Z(n)$ due to Theorem \ref{th: MLN_Normal_Form}.

\begin{proposition}
    \label{cor: partition}
    Let $\Phi$ be an MLN in the form \eqref{eq:mln_normal_form}, then the partition function $Z(n)$ is given as:
\begin{equation}
    Z(n) = \sum_{\bk}\binom{n}{\bk}\prod_{i\in [u]}s_i^{k_i} \prod_{\substack{{i,j \in [u]}\\{i\leq j}}}(f_{ij})^{\bk(i,j)} 
\end{equation}
where $\bk(i,j)$ is defined in equation \eqref{kij}.
\end{proposition}
\begin{proof}[Sketch]
The proposition is a parameterized version of equation \eqref{eq:fomc-universal-form}, where $\prod_{i\in [u]}s_i^{k_i}$ takes into account the weight contributions due to the 1-type realizations and  $f_{ij}$ is essentially a weighted version of $n_{ij}$ i.e. given a pair of constants $c$ and $d$ such that they realize the $i^{th}$ and the $j^{th}$ 1-type respectively, then $f_{ij}$ is the sum of the weights due to the 2-types realized by the extensions to the binary predicates containing both $c$ and $d$.
  
\end{proof}

    



%% file: 05_projectivity.tex
\section{Projectivity in Markov Logic Networks }
In this section we present  the necessary and sufficient conditions for an MLN to be projective in the two variable fragment. We also show that the provided characterization  is strictly more expressive than the characterization provided by Jaeger and Schulte \cite{Projectivity_first}.

   \begin{lemma} [Sufficiency]
    \label{lem: sufficiency_factorisation}
    A Markov Logic Network in the two variable fragment
    is projective if: 
    \begin{equation}
      \forall i,j,i'j' \in  [u]: f_{ij} = f_{i'j'} 
    \end{equation}
  \end{lemma}
\begin{proof}
  Let $\forall i,j,i'j' \in  [u]: f_{ij} = f_{i'j'} = S$. Hence, due to Proposition~\ref{cor: partition}, we have:
  \begin{align}
    \label{eq: partition_factorises}
    Z(n) &= \sum_{\bk}\binom{n}{\bk}\prod_{i\in [u]}s_i^{k_i} \prod_{\substack{{i,j \in [u]}\\{i\leq j}}}(S)^{\bk(i,j)} \\
    &= \sum_{\bk}\binom{n}{\bk}\prod_{i\in [u]}s_i^{k_i} (S)^{\binom{n}{2}} = S^{\binom{n}{2}}\Big(\sum_{i\in[u]}s_i \Big)^{n}
  \end{align}
  Let $p_i = \Bigl(\frac{s_i}{\sum_{i}s_i}\Bigr)$ and $w_{ijl} = \Bigl(\frac{t_{ijl}}{S}\Bigl) $. Hence,
  \begin{align*}
      P^{(n)}_{\Phi}(\omega) &= \frac{1}{(\sum_{i\in[u]}s_i )^{n}(S)^{\binom{n}{2}}} \prod_{i \in [u]} s_{i}^{k_i} \prod_{\substack{i,j\in [u]\\ i \leq j}}\prod_{l \in [b]} (t_{ijl})^{h^{ij}_l}\\
      &= \prod_{i \in [u]} \Bigl(\frac{s_{i}}{\sum_{i\in[u]}s_i}\Bigr)^{k_i} \prod_{\substack{i,j\in [u]\\ i \leq j}}\prod_{l \in [b]} \bigl(\frac{t_{ijl}}{S}\bigr)^{h^{ij}_l}\\
      & = \prod_{i \in [u]} p_{i}^{k_i} \prod_{\substack{{i,j \in [u]} \\{i \leq j } }}\prod_{l\in[b]} w_{ijl}^{h^{ij}_l}
  \end{align*}
Using the multi-relational representation, $P^{(n)}_{\Phi}(\omega)$ can be equivalently expressed as:
\begin{align}
  \label{eq: RBM}
  P^{(n)}_{\Phi}(\bX=\x,\bY = \y) &= \prod_{q \in [n]} p_{x_q} \prod_{\substack{{q,r \in [n]} \\{q <r } }} w_{x_qx_ry_{qr}} 
\end{align}  
Let $(\bX',\bY')$ be the random vector containing $X_q$ and $Y_{p,q}$
with $p<q\in[m]$.
Clearly, our goal is to show that $$P^{(n)}_{\Phi}\downarrow [m] (\bX' = \x',\bY' = \y') = P^{(m)}_{\Phi}(\bX' = \x',\bY' = \y')$$
Now, the marginal distribution over the $m$-worlds $(\bX',\bY')$, due to $P^{(n)}_{\Phi}(\bX=\x,\bY = \y)$ can be expressed as: 
    \begin{align*}
      &P^{(n)}_{\Phi}\downarrow [m] (\bX' = \x',\bY' = \y') = \sum_{\substack{\x_{[m]} = \x' \\\y_{[m]} = \y'}} P^{(n)}_{\Phi}(\bX = \x,\bY = \y)\\
                                 &= \sum_{\substack{\x_{[m]} = \x' \\\y_{[m]} = \y'}}\prod_{q \in [n]} p_{x_q} \prod_{\substack{{q,r \in [n]} \\{q <r } }} w_{x_qx_ry_{qr}}\\
                                 &= \prod_{q \in [m]} p_{x_q} \prod_{ \substack{{q,r \in [m]} \\{q <r } }} w_{x_qx_ry_{qr}} \times \Biggl(\sum_{\substack{\x_{[\overline{m}]} \\\y_{[\overline{m}]}}} \prod_{q \in [\overline{m}]} p_{x_q} \prod_{ \substack{{q,r \in [\overline{m}]} \\{q <r } }} w_{x_qx_ry_{qr}}  \prod_{\substack{q \in [m]\\  r \in [\overline{m}]  }} w_{x_qx_ry_{qr}}  \Biggr)\\
                                 &= \prod_{i \in [u]} p_i^{k_i(\bx')} \prod_{\substack{{i,j \in [u]} \\ {i\leq j}}} \prod_{l \in [b]} w_{ijl}^{h^{ij}_l(\y')}\\
                                 &\times \Biggl(\sum_{\substack{\x_{[\overline{m}]} \\\y_{[\overline{m}]}}} \prod_{i \in [u]} p_{i}^{k_i(\x_{[\overline{m}]})} \prod_{\substack{{i,j \in [u]} \\ {i \leq j}}  } \prod_{l \in [b]} w_{ijl}^{h^{ij}_l (\y_{[\overline{m}]})} \prod_{\substack{i,j\in [u] \\ i \leq j }} \prod_{l \in [b]}w_{ijl}^{h^{ij}_l({{[m] \otimes [\overline{m}]}})} \Biggr)\\
    \end{align*}
  where $A\otimes B = A\times B \cup B\times A$. 
    Notice that $\prod_{i \in [u]} p_i^{k_i(\bx')} \prod_{\substack{{i,j \in [u]} \\ {i\leq j}}} \prod_{l \in [b]} w_{ijl}^{h^{ij}_l(\y')}$ is $P^{(m)}_{\Phi}(\bX'=\x',\bY'=\y')$. Hence, in order to complete the proof, we will now show that  for any $\x'$: 
    \begin{equation}
      \label{eq: increment_sufficiency}  
    \sum_{\substack{\x_{[\overline{m}]} \\\y_{[\overline{m}]}}} \prod_{i \in [u]} p_{i}^{k_i(\x_{[\overline{m}]})} \prod_{\substack{{i,j \in [u]} \\ {i \leq j}}  } \prod_{l \in [b]} w_{ijl}^{h^{ij}_l (\y_{[\overline{m}]})} \prod_{\substack{i,j\in [u]\\ i \leq j}} \prod_{l \in [b]}w_{ijl}^{h^{ij}_l({{[m] \otimes [\overline{m}]}})} = 1
  \end{equation}
   The LHS of equation \eqref{eq: increment_sufficiency} can be written as:
    \begin{align} 
      \label{eq: increment_2}
        \sum_{\sum{\bk} = n-m} \binom{n-m}{\bk}\prod_{i \in [u]} p_{i}^{k_i} \prod_{\substack{{i,j \in [u]} \\{ i \leq j } }} \bigl(\sum_{l}w_{ijl}\bigr)^{\bk(i,j)}\prod_{\substack{{i,j \in [u]} \\ {i\leq j}}} \bigl(\sum_{l} w_{ijl}\bigr)^{k_{i}(\x') \times k_j}
    \end{align}
    By definition, for any $i,j \in [u]$, $\sum_{l} w_{ijl} = 1$ , and $\sum_{i}p_i = 1$. Hence, expression \eqref{eq: increment_2} can be written as:
    \begin{align*}
          \sum_{\sum \bk = n-m}  \binom{n-m}{\bk}\prod_{i \in [u]} p_{i}^{k_i} = \Biggl(\sum_{i} p_{i}\Biggr)^{n-m} = 1
    \end{align*}
    Hence, completing the proof.
  \end{proof}
  
  We will now prove that the aforementioned sufficient conditions are also necessary. 
  
  \begin{lemma}[Necessary]
    \label{lem: neccesity}
    If a Markov Logic network in the two variable fragment is projective then :
    \begin{equation}
      \forall i,j,i',j'\in [u]: f_{ij} =  f_{i'j'}  
    \end{equation}
    
  \end{lemma}
  \begin{proof} Let us have a markov logic network $\Phi$ over a domain $[n+1]$. Let $\bX'$ and $\bY'$ be random variables representing multi-relational graphs on the domain $[n]$. Then :
  
  \begin{align*}
      &P_{\Phi}^{(n+1)}\downarrow [n] (\bX' = \x',\bY' = \y') = \sum_{\substack{\x_{[n]} = \x' \\\y_{[n]} = \y'}} P_{\Phi}^{(n+1)}(\bX = \x,\bY = \y)\\
                               &= \sum_{\substack{\x_{[n]} = \x' \\\y_{[n]} = \y'}}\frac{1}{Z(n+1)}\prod_{q\in [n+1]} s_{x_q} \prod_{\substack{{q,r \in [n+1]} \\ {q < r}}} t_{x_qx_ry_{qr}} \\
                               &= \frac{1}{Z(n)}\prod_{q\in [n]} s_{x_q} \prod_{\substack{{q,r \in [n]} \\ {q < r}}} t_{x_qx_ry_{qr}} \frac{Z(n)}{Z(n+1)} \sum_{\substack{x_{n+1}\\ y_{q,n+1}}}s_{x_{n+1}} \prod_{q\in [n]}t_{x_qx_{n+1}y_{q,n+1}} \\
                               &=  P^{(n)}_{\Phi}(\bX' = \x',\bY' = \y') \frac{Z(n)}{Z(n+1)} \sum_{\substack{x_{n+1}\\ y_{q,n+1}}}s_{x_{n+1}} \prod_{q\in [n]}t_{x_qx_{n+1}y_{q,n+1}} \\ 
  \end{align*}
  Due to projectivity we have that:
  \begin{align*}
    P^{(n)}_{\Phi}(\bX' = \x',\bY' = \y') &= P^{(n+1)}_{\Phi}\downarrow [n] (\bX' = \x',\bY' = \y') \\  
  \end{align*}
  Hence,
  \begin{align*}
    \frac{Z(n+1)}{Z(n)} &=  \sum_{\substack{x_{n+1} \\ y_{q,n+1}}}s_{x_{n+1}} \prod_{q\in [n]}t_{x_qx_{n+1}y_{q,n+1}}
  \end{align*}
  which can be equivalently written as:
  \begin{align*}  
    \frac{Z(n+1)}{Z(n)} &= \sum_{i\in [u]}s_i\prod_{j\in[u]}\Bigr(\sum_{l \in [b]}  t_{jil}\Bigl)^{k_{j}(\bx')} 
  \end{align*}
  Now, $\sum_{l \in [b]}  t_{jiv} = f_{ji} = f_{ij}$. Hence:
  \begin{align*}
    \frac{Z(n+1)}{Z(n)} &= \sum_{i\in [u]}s_i \prod_{j\in[u]}f_{ij}^{k_{j}(\bx')} 
  \end{align*}
  Hence, for any choice of the domain size $m$ and for any choice of $m$-worlds $(\x,\y)$ and $(\x',\y')$, we have that:
  \begin{align}
    \sum_{i\in [u]}s_i \prod_{j\in[u]}f_{ij}^{k_{j}(\bx)} &= \sum_{i\in [u]}s_i \prod_{j\in[u]}f_{ij}^{k_{j}(\bx')} 
  \end{align}
which implies \footnote{We prove this \say{implication} in detail in the Appendix.} that:
\begin{equation*}
      \forall i,j,i',j'\in [u]: f_{ij} =  f_{i'j'}  
\end{equation*}
Hence, completing the proof.
\end{proof}
  We are finally able to provide the following theorem.
  
  \begin{theorem}
    \label{th: MLN_Projectivity}
    A Markov Logic Network (MLN) $\Phi = \{(\phi_i,a_i)\}_i$, such that $\phi_i$ contains at-most two variables is projective if and only if,
    \begin{equation} 
      \label{eq: main_result} 
    \forall i,j,i',j' \in [u] : f_{ij} = f_{i'j'} 
    \end{equation}
  
  \end{theorem} 

  In the next section, we will show that the conditions in
  Theorem \ref{th: MLN_Projectivity} correspond to a special type of
  probability distributions. We will characterize such distributions
  and  then investigate their various aspects.

%% file: 06_rbm.tex
\section{Relational block model}

In this section we introduce the Relational Block Model (RBM). We show that any projective MLN in the two variable fragment can be expressed as an RBM. Furthermore, an RBM is a 
unique characterization of any projective MLN.
\begin{definition}
  \label{def: RBM}
  Let $n$ be a positive integer (the number of domain constants), $u$ be a
  positive integer (the number of 1-types), $b$ be a positive integer (the number of 2-tables), $p = (p_1,...,p_u)$ be a probability vector on $[u]=\{1,...,u\}$ 
  and $W  = (w_{ijl}) \in [0,1]^{u\times u \times b}$, where $w_{ijl} = w_{\underline{ijl}}$  ($w_{ijl}$ is the conditional probability of domain elements $(c,d)$ realizing the $l^{th}$ 2-table, given $ {i(c)}$ and $ {j(d)}$).
  The multi-relational graph $(\x,\y)$ is drawn under $\mathrm{RBM}(n,p,W)$ if $\x$ is an $n$-dimensional vector with $i.i.d$ components distributed under $p$ and $\y$ is a random vector with its component $y_{qr} = l$, where $l \in [b]$, with a 
  probability $w_{x_qx_rl}$ independently of all other pair of domain constants. 
  
  
  Thus, the probability distribution of $(\x,\y)$ is defined as follows, where $\x \in [u]^{n}$ and $\y \in [b]^{\binom{n}{2}}$ 
\begin{align*}
  P(\X=\x) &:= \prod_{q=1}^{n} p_{x_q} = \prod_{i=1}^{u} p_{x_i}^{k_i}\\
  P(\Y=\y|\X = \x) &:= \prod_{1\leq q < r \leq n  } w_{x_qx_ry_{pq}} \\
  &= \prod_{ 1\leq i \leq j \leq u} \, \prod_{1\leq l \leq b}(w_{ijl})^{h^{ij}_l}
\end{align*}
 
\end{definition}



In the following example, we show how RBMs can model homophily. 

\begin{example}[Homophily] Let us have an FO$^2$ language with a unary predicate $C$ (representing a two colors) and a binary predicate $R$. 
  We wish to model a distribution on simple undirected graphs i.e. models of the formula $\phi = \forall xy. \neg R(x,x) \land (R(x,y) \rightarrow R(y,x))$ such that 
  same color nodes are more likely to have an edge. Due to $\phi$ the 1-types with $\neg R(x,x)$ as a conjunct have a probability zero. Hence, we can assume we have only two 1-types:
  $1(x) = C(x)\land \neg R(x,x) $ and $2(x) = \neg C(x)\land \neg R(x,x) $ (representing two possible colors for a given node). Similarly due to $\phi$, we have only two 2-tables $1(x,y): R(x,y) \land R(y,x)$ and $ 2(x,y): \neg R(x,y) \land \neg R(y,x)$ (representing existence and non existence of edges). We can now easily define homophily by following parameterization of an RBM.
  $p_1 = p_2 = 0.5$ i.e. any node can have  two colors with equal probability. Then we can define $w_{111} = 0.9$, $w_{112} = 0.1$, $w_{221} = 0.9$, $w_{222} = 0.1$, $w_{121} = 0.1$ and $w_{122}=0.9$. 
\end{example}

\begin{theorem}
  \label{th:RBM_MLN}
  Every projective Markov Logic Network in the two variable fragment can be expressed as an RBM.
\end{theorem}
\begin{proof}
  The proof follows from the sufficiency proof in Lemma \ref{lem: sufficiency_factorisation}. Notice that in the proof, we derive equation \eqref{eq: RBM}, which is exactly the expression for RBM. Hence, any projective MLN can be converted to an RBM by defining $p_i$ and $w_{ijl}$ as follows:
  \begin{align}
    \label{eq: MLN_2_RBM}
    p_{i} = \frac{s_i}{\sum_{i}s_i} &\quad w_{ijl} = \frac{t_{ijl}}{\sum_{l}t_{ijl}}
  \end{align}
\end{proof}

\begin{theorem}
  \label{th: MLN_RBM}
  Every RBM can be expressed as a projective MLN in the two variable fragment. 
\end{theorem}
\begin{proof}
  Given an RBM as defined in Definition \ref{def: RBM}, let us have an MLN $\Phi$ such that every 1-type $i(x)$ is a formula in the MLN with a weight $\log{p_i}$. $\Phi$ also has a weighted formula $ijl(x,y)$ for every 2-type, such that $i \leq j$. The weight for $ijl(x,y)$ is $\log(w_{ijl})$ if $ijl(x,y) \neq ijl(y,x)$, and is 
  $0.5\log(w_{ijl})$ if $ijl(x,y) = ijl(y,x)$. It can be seen from definition of  $s_i$ \eqref{si} and $t_{ijl}$ \eqref{t_ijl} that $s_i = p_i$ and $t_{ijl} = w_{ijl}$. Hence, due to \eqref{eq:mln_normal_form}, we have that:
  
  \begin{equation}
P^{(n)}_{\Phi}(\omega) = \frac{1}{Z(n)} \prod_{i \in [u]} p_{i}^{k_i} \prod_{\substack{i,j\in [u]\\ i \leq j}}\prod_{l \in [b]} (w_{ijl})^{h^{ij}_l}
\end{equation}
Now, $\sum_{i} p_i = 1$ and $\sum_{l} w_{ijl} = 1$. Hence, using Proposition \ref{cor: partition}, we have that $Z(n)=1$. Hence, completing the proof.

\end{proof}

\begin{proposition}
  \label{th: RBM_unique}
  Given two RBMs with probability distribution $P'$ and $P''$ and parameters $\{p'_i,w'_{ijl}\}$ and $\{p''_i,w''_{ijl}\}$. If $P'=P''$, then:
  \begin{align*}
    p'_i = p''_i \quad w'_{ijl} = w''_{ijl}
  \end{align*} 
\end{proposition}
\begin{proof}

The proposition is a consequence of the fact that the parameter $p_i$ is marginal probability of an arbitrary constant $c$ realizing the $i^{th}$ 1-type and $w_{ijl}$ is the conditional probability of an arbitrary pair of constants $(c,d)$ realizing the $l^{th}$ 2-table given $i(c)$ and $j(d)$. Hence, two distributions that disagree on the $p_i$ and $w_{ijl}$  cannot assign the same probability to marginal probability of $i(c)$ and $ijl(c,d)$ and hence, cannot be the same distribution.

\end{proof}

\begin{corollary}[of Proposition \ref{th: RBM_unique}]
  \label{cor: MLN_unique}
  Given two MLNs $\Phi'$ and $\Phi''$ such that they have the same probability distributions $P_{\Phi'}$ and $P_{\Phi''}$, with there respective RBMs parameterized by $\{p'_i,w'_{ijl}\}$ and $\{p''_i,w''_{ijl}\}$. Then we must have that:
  \begin{align*}
    p'_i = p''_i \quad w'_{ijl} = w''_{ijl}
  \end{align*} 
\end{corollary}

Hence, RBMs are a unique representation for projective MLNs in the two variable fragment.

%% file: 07_comparison.tex
\section{Comparison to Previous Characterizations of Projectivity }

Jaeger and Schulte \cite{Projectivity_first} show that an MLN is projective if it's formulae $\phi_i$ satisfy the property that any two atoms appearing in $\phi_i$ contain exactly the same variables. Such MLNs are also known as \emph{$\sigma$-determinate} \cite{MLN_inf}. We now show that in the two variable fragment, theorem \ref{th: MLN_Projectivity}  leads to a strictly more expressive class of MLNs.

\begin{proposition}
  \label{prop: more_than_manfred}
Given an MLN $\Phi = \{\phi_i,a_i\}_i$ such that any two atoms appearing in $\phi_i$ contain exactly the same variables or equivalently that the MLN is $\sigma-$determinate. Then:
\begin{equation}
  \forall i,j,i',j' \in [u], \forall l \in [b]: t_{i'j'l} = t_{ijl} 
\end{equation} 
\end{proposition}

\begin{proof}[Sketch]
  We first write an equivalent MLN $\Phi' = \{\alpha_q(x),a_q\} \cup \{\beta_p(x,y),b_p\} $ as presented in proof of theorem \ref{th: MLN_Normal_Form}. Due to the conditions provided in the proposition, all the atoms in $\beta_p(x,y)$  contain both the variables $x$ and $y$. Using the definition of $t_{ijl}$ from \eqref{t_ijl}, and the fact that none of the $\beta_p(x,y)$ have an atom with only one variable, we have that the value of $t_{ijl}$ depends only on the $l^{th}$ 2-table, irrespective of the 1-types $i$ and $j$. This is because, none of the first order atoms in the $i^{th}$ and the $j^{th}$ 1-type appear in $\beta_{p}(x,y)$. Hence, $t_{ijl}$ only depends on $l$. Hence:
  $$\forall i,j,i',j' \in [u], \forall l \in [b]: t_{i'j'l} = t_{ijl}   $$
  
\end{proof}
Proposition \ref{prop: more_than_manfred} is a stricter condition than theorem \ref{th: MLN_Projectivity}. In the following, we prove that $\sigma$-determinate MLNs cannot express all the projective MLNs in the two variable fragment.

\begin{proposition}
 There exists a projective MLN in the two variable fragment which cannot be expressed as  a $\sigma-$determinate MLN.
\end{proposition}
\begin{proof}
  Let us have a $\sigma$-determinate MLN $\Phi$, since $\Phi$ is projective, we can create it's equivalent RBM (due to Theorem \ref{th: MLN_RBM}), say $P$.  Let $\{p_i,w_{ijl}\}$ be the parameters of $P$. Due to equation \eqref{eq: MLN_2_RBM} and proposition \ref{prop: more_than_manfred}, we have that $w_{ijl} = w_{i'j'l}$ for all $i,j,i',j'$. Due to existence of an MLN for every RBM (from theorem \ref{th: MLN_RBM}), we can always create an MLN $\Phi'$ for which the RBM parameters $w_{ijl} \neq w_{i'j'l}$ for some $i,j,i',j'$. Since,  RBMs uniquely characterize the probability distributions due to MLNs {(from corollary \ref{cor: MLN_unique})}, $\Phi'$ can not be expressed as an MLN such that  $w_{ijl} = w_{i'j'l}$. Hence, $\Phi'$ can not be expressed as a  $\sigma$-determinate MLN. 
  
\end{proof}

%% file: 08_learning.tex
\section{Maximum Likelihood Learning}
In a learning setting, for an MLN $\{\phi_i,a_i\}$ in the two variable fragment, we are interested in estimating the set of parameters $\bm{\theta} = \{a_i\}$
that maximize the likelihood of a training example such that the learnt MLN is projective. As analyzed in \cite{one_Network,Relational_Marginal_Polytope}, we will focus 
on the scenario where only a single possible world $\omega \in \Omega^{(n)}$ is observed. We estimate $\bm{\theta}$ by maximizing the likelihood
\begin{equation}
    \label{likelihood}
    L^{(n)}(\bm{\theta}|\omega) = P^{(n)}_{\bm{\theta}}(\omega)
\end{equation}

Notice that although every projective MLN can be equivalently defined as an RBM. The maximum likelihood  parameter estimate for an RBM is not the same as the  parameter estimate for an MLN such that it is projective. 

We will now provide, the maximum likelihood estimator for an RBM. This estimator is completely analytically definable and admits many desirable consistency and efficiency properties. We will then show that, as far as maximizing data likelihood is concerned, RBM is at least as good as any projective MLN.

\begin{proposition}
    \label{lem: RBM_ML}
    Given a training example $\omega \in \Omega^{(n)}$, the maximum likelihood parameter estimate for an RBM is given as :
    \begin{align}
        p_i = \frac{k_i}{n}    
        & &w_{ijl} = \frac{h^{ij}_l}{\bk(i,j)}
    \end{align}
\end{proposition}
Proposition \ref{lem: RBM_ML} can be derived by maximizing the log likelihood due to the distribution given in Definition \ref{def: RBM}.

We will now see how maximum likelihood parameter estimate can be obtained for an MLN such that the MLN is projective. 

Given an MLN $\{\phi_i,a_i\}_i$ in the two variable fragment, where $\bm{\theta} = \{a_i\}_i$ are unknown parameters to be estimated, due to Theorem \ref{th: MLN_Normal_Form}, we can define $s_i(\bm{\theta})$ and $t_{ijl}(\bm{\theta})$, such that the likelihood is given as:

\begin{equation}
L(\omega|\bm{\theta}) = \frac{1}{Z(n)} \prod_{i \in [u]} s_{i}(\bm{\theta})^{k_i} \prod_{\substack{i,j\in [u]\\ i \leq j}}\prod_{l \in [b]} (t_{ijl}(\bm{\theta}))^{h^{ij}_l}
\end{equation}
The maximum likelihood parameter estimates such that the estimated MLN is projective,  can be then obtained by solving the following optimization problem:
\begin{align}
    \label{MLN_ML_proj}
    \begin{split}
        \underset{\theta}{maximize}:  &\Big[ \sum_{i\in [u]}k_i \log s_{i}(\bm{\theta}) + \sum_{\substack{i,j \in [u] \\ i \leq j }}\sum_{l\in [b]}h^{ij}_l\log{t_{ijl}(\bm{\theta})} \\
        & - n\log\big( \sum_{i\in [u]} s_{i}(\bm{\theta}) \big) - \binom{n}{2} \log f_{11} \Big]\\
        subject \quad to:   &\forall i,j,i',j' \in [u] : f_{ij} = f_{i'j'}         
    \end{split}
\end{align}
Notice, that $- n\log\big( \sum_{i\in [u]} s_{i}(\bm{\theta}) \big) - \binom{n}{2} \log f_{11}$ represents $-\log(Z(n)$ and $f_{11}$ could have been any particular $f_{ij}$ (see \eqref{eq: partition_factorises}). The above optimization can be solved through any conventional optimization algorithm. It can be seen that this problem has a much lesser overhead as far as computing $\log(Z(n))$ is concerned. But the additional constraints may counter act this gain. 


\begin{proposition}
    \label{prop: RBM_best}
    Given a training example $\omega \in \Omega^{(n)}$, then there is no parameterization for any projective MLN in the two variable fragment that has a higher likelihood for 
    $\omega$ than the maximum likelihood RBM for $\omega$.
\end{proposition}
\begin{proof}
    Let $L$ be the likelihood of $\omega$ due to the maximum likelihood RBM. Let $L'$ be the likelihood of $\omega$ due to a projective MLN $\Phi$, such that $L' > L$. Now, due to Theorem \ref{th:RBM_MLN}, $\Phi$ can be expressed as an RBM. Hence, we can have an RBM such that the likelihood of $\omega$ is $L'$, but $L' > L$ which is a contradiction. Hence, we cannot have a projective MLN that gives a higher likelihood to $\omega$ than the maximum likelihood RBM.
\end{proof}
Proposition \ref{prop: RBM_best} shows us that if a data source is known to be projective (i.e. we know that marginals in the data will be independent of the domain at large) then in terms of likelihood, specially in the case of large relational datasets, we are better off in using an RBM than an expert defined MLN. This can also be argued 
from efficiency point of view as RBMs admit much more efficient parameter estimates.

We will now move on to the question: \emph{are  parameters learned on a domain of size $n$, also good for modelling  domain of a different size $m$ ?} This question is an abstraction of 
many real world problems, for example, learning over relational data in presence of incomplete information \cite{social_Network_missing}, modelling a social network from only  sub-sampled populations \cite{social_Network_sample}, modelling progression of a
disease in a population by only testing a small set of individuals \cite{test_disease} etc. 

Jaeger and Schulte \cite{Projectivity_first} formalized the afore mentioned notions in the following two criterions:

\begin{align}
    \label{eq:consistency_1}
    E_{\omega}[\argmax_{\bm{\theta}} \log L^{(m)}(\bm{\theta}|\omega')] &= \argmax_{\bm{\theta}} \log L^{(n)}(\bm{\theta}|\omega) 
\end{align}
\begin{align}
\label{eq: consistency_2}
    \argmax_{\bm{\theta}} E_{\omega}[\log L^{(m)}(\bm{\theta}|\omega')] &= \argmax_{\bm{\theta}} \log L^{(n)}(\bm{\theta}|\omega)
\end{align}

It is easy to see, by law of large numbers, that RBMs satisfy both these criterions. On the other hand the same can not be said about the maximum likelihood estimates for projective MLNs as described in  \eqref{MLN_ML_proj}. 

%% file: 09_conclusion.tex
\section{Conclusion}
In this work, we have characterized the class of projective MLN in the two-variable fragment. We have also recognized a special MLN amongst such MLNs, namely Relational Block Model. We show that the maximum likelihood RBM maximizes the training data likelihood 
w.r.t to any projective MLN in the two-variable fragment. Furthermore, RBMs admit consistent parameter learning from sub-sampled domains, potentially allowing them to scale to very large datasets, especially in situations where the test data size is not known or changes over time. 

From an applications point of view, the superiority of RBMs in terms of training likelihood maximization and consistent parameter learning can potentially make them a better choice over an expert defined MLN, especially when training set is large and the test domain size is unknown or varies over time. We plan to investigate such capabilities of RBMs and Projective MLNs in future work, especially in comparison to models like Adaptive MLNs \cite{A_MLN} and Domain Size Aware MLNs \cite{DA_MLN}.

On the theoretical front, the imposed independence structure due to projectivity clearly resembles the AHK models proposed in \cite{Projectivity_SRL}. In future works, we aim at investigating this resemblance and generalizing our work to capture complete projectivity criterion for all the MLNs.

\section{Acknowledgements}
We would like to thank Manfred Jaeger and Felix Weitkämper for their valuable critique and discussion time on the topic.

%% file: appendix.tex
\section*{Appendix}
\subsection*{ A.1 : Lemma \ref{lem: neccesity} [Necessary]}
In this section we will make the last steps in the proof of lemma \ref{lem: neccesity} more rigorous. In the lemma we argue that, for any choice of the domain size $m$ and for any choice of $m$-worlds $(\x,\y)$ and $(\x',\y')$, we have that:
  \begin{align}
    \label{eq: lemma_2_condition}
    \sum_{i\in [u]}s_i \prod_{j\in[u]}f_{ij}^{k_{j}(\bx)} &= \sum_{i\in [u]}s_i \prod_{j\in[u]}f_{ij}^{k_{j}(\bx')} 
  \end{align}
This implies that:
\begin{equation}
    \label{eq: lem_2_conclusion}
      \forall i,j,i',j'\in [u]: f_{ij} =  f_{i'j'}  
\end{equation}
We will first infer a slightly stricter equation from \eqref{eq: lemma_2_condition}.
Since, $f_{ij} = f_{ji}$, we can see $\{f_{ij}\}$ as a symmetric $u \times u$ matrix in $\mathbb{R}_{>0}^{u\times u}$. Furthermore, $\x$ and $\x'$ can have any
1-type cardinalities $\bk = \langle k_1 ... k_u\rangle$ and $\bk'= \langle k'_1 ... k'_u\rangle$ respectively, such that $\sum_{i\in [u]}k_i = \sum_{i\in [u]}k'_i = m$. Hence, we can conclude that, for all $\bk$ and $ \bk'$ such that $\sum_{i\in [u]}k_i = \sum_{i\in [u]}k'_i$, we have that:
\begin{align}
    \label{eq: lemma_2_condition_2}
    \sum_{i\in [u]}s_i \prod_{j\in[u]}f_{ij}^{k_j} &= \sum_{i\in [u]}s_i \prod_{j\in[u]}f_{ij}^{k'_{j}} 
  \end{align}
Hence, our goal is to prove that \eqref{eq: lemma_2_condition_2} implies \eqref{eq: lem_2_conclusion}. We formally prove this statement in Lemma \ref{lemma_2}. Before proving Lemma \ref{lemma_2}, we will need to prove the following 
auxiliary lemma. 
\begin{lemma}
    \label{lemma_1}
  Let $(x_i)^{m}_{i=1}$,$(y_i)^{m}_{i=1}$ and $(a_i)^{m}_{i=1}$ be tuples of positive non-zero reals. If for all positive integers $n$: 
    \begin{equation}
      \label{eq: lemma_1}
     \sum_{i=1}^{m}a_ix_i^{n} = \sum_{i=1}^{m}a_iy_i^{n}
    \end{equation}
  then the set of entries in $(x_i)^{m}_{i=1}$ and the set of entries in $(y_i)^{m}_{i=1}$ are the same. 
  \end{lemma}
    
  \begin{proof}Let $\{u_{i}\}^{p}_{i=1}$ and  $\{v_{i}\}^{q}_{i=1}$ be the set of unique entries in $(x_i)^{m}_{i=1}$ and $(y_i)^{m}_{i=1}$ respectively. Also, without loss of generality, we may assume an ordering such that $u_1 > u_2> ... >u_{p} $ and $v_1 > v_2> ...>v_{q} $ and also that $q\geq p $. We can rewrite \eqref{eq: lemma_1} as:
  \begin{equation}
    \label{eq: lemma_1_equivalence}
   \forall n \in \mathbb{Z^{+}}:  \sum_{i=1}^{p}c_iu_i^{n} = \sum_{i=1}^{q}d_iv_i^{n} 
  \end{equation}
  As $n$ grows the leading term on LHS is $c_1u_{1}^{n}$ and on the RHS is $d_1v_{1}^{n}$. Hence, it must be :
  
  \begin{equation*}
    \forall n \in \mathbb{Z^{+}}: c_1 u_{1}^{n}  = d_1 v_{1}^{n} 
  \end{equation*}
  Since, $u_1,v_1,c_1$ and $d_1$ are non-zero positive reals, we can conclude that $u_1=v_1$ and $c_1 = d_1$. Hence, we may subtract $c_1 u_{1}^{n}$ from both sides in \eqref{eq: lemma_1_equivalence} to get :
  \begin{equation}
    \label{eq: lemma_1_equivalence_1}
   \forall n \in \mathbb{Z^{+}}:  \sum_{i=2}^{m'}c_iu_i^{n} = \sum_{i=2}^{m''}d_iv_i^{n} 
  \end{equation}
  We may now repeat the aforementioned argument and infer that $u_2=v_2$ and $c_2 = d_2$. Furthermore, repeating this argument $p$ times, we can infer that $\{u_i\}^{p}_{i=1} = \{v_i\}^{p}_{i=1}$, leaving us with $0 =\sum_{i=q-p+1}^{p}d_iv_i^{n}$, which is a contradiction, hence, $p=q$. Hence, we have that $\{u_{i}\}^{p}_{i=1}$ = $\{v_{i}\}^{q}_{i=1}$. Hence, completing the proof.
  \end{proof}

  \begin{lemma}
    \label{lemma_2}
  Let $F = (f_{ij}) \in \mathbb{R}_{>0}^{u \times u}$ be a symmetric matrix and let $(s_i)^{u}_{i=1} \in \mathbb{R}_{>0}^{u}$. If for all $\bm{k} = \langle k_1,...,k_u \rangle$ and $\bm{k'} = \langle k'_1,...,k'_u \rangle $ such that $k_i,k'_i \in \mathbb{Z^{+}}$ and $\sum_{i=1}^{u}k_i = \sum_{i=1}^{u}k'_i$, we have that: 
    \begin{equation}
      \label{eq: lemma_2}
  \sum_{i=1}^{u}s_i\prod_{j\in [u]}f_{ij}^{k_{j}} = \sum_{i=1}^{u}s_i\prod_{j\in [u]}f_{ij}^{k'_{j}}
    \end{equation}
  then 
  \begin{equation*}
    \forall i,j,i',j' : f_{ij} = f_{i'j'}
  \end{equation*}
  \end{lemma}
  
  \begin{proof}
  Let $\bm{k}$ be such that $k_p=n$, let $k_i = 0$ for all $i \neq p$. Let $\bm{k'}$ be such  that $k_q=n$ and $k_i = 0$ for all $i \neq q$. Then due to \eqref{eq: lemma_2}, we have that:
  
  \begin{equation}
  \forall n \in \mathbb{Z^{+}}: \sum_{i=1}^{u}s_i(f_{ip})^{n} = \sum_{i=1}^{u}s_i(f_{iq})^{n}
  \end{equation} 
  Hence, due to Lemma \ref{lemma_1}, we have that the entries in $(f_{ip})^{u}_{i=1}$ and $(f_{iq})^{u}_{i=1}$ form the same set. A similar argument can 
  be repeated for any pair of columns. Hence, all columns in $F$ have the same set of entries, we denote the set of such entries as $U$. 
  
  Now, let $n = uk$ where $k \in \mathbb{Z}^{+}$, and $\bm{k}$ such that $k_i=k$ for all $i \in [u]$ and $\bm{k'}$ such that $k'_q=n$ and $k'_i = 0$ for all $i \neq q$. Then due to \eqref{eq: lemma_2}, we have that:
  \begin{align*}
    \forall k \in \mathbb{Z^{+}}: \sum_{i=1}^{u} s_i \prod_{p\in[u]}f_{ip}^{k} &= \sum_{i=1}^{u}s_i(f_{iq})^{uk}\\
    \forall k \in \mathbb{Z^{+}}: \sum_{i=1}^{u} s_i \bigl(\prod_{p\in[u]}f_{ip}\bigr)^{k} &= \sum_{i=1}^{u}s_i (f_{iq}^{u}\bigr)^{k}
  \end{align*} 
  As $k$ grows the leading term on left hand side and right hand side must agree for the equality to hold. Let $c_{i'} \bigl(\prod_{p\in[u]}f_{i'p}\bigr)^{k}$ and $d_{i''} (f_{i''q}^{u}\bigr)^{k}$ be the leading terms on RHS and LHS respectively.
  Hence,
  \begin{equation}
    \label{eq: compare_uni_edge}
    \forall k \in \mathbb{Z^{+}} : c_{i'} \bigl(\prod_{p\in[u]}f_{i'p}\bigr)^{k} = d_{i''} (f_{i''q}^{u}\bigr)^{k} 
  \end{equation}
  which implies that $\prod_{p\in[u]}f_{i'p} = f_{i''q}^{u}$. Now, clearly $f_{i''q}$ is equal to the maximum term in $U$ say $s$. Now, $\prod_{p\in[u]}f_{i'p}$ is a product of all  the terms in 
  the $p^{th}$ matrix column of $F$. Hence, $\prod_{p\in[u]}f_{i'p} \leq s^{u}$. Hence, due to \eqref{eq: compare_uni_edge}, we have that:
  
  \begin{equation*}
    \forall i,j,i',j' : f_{ij} = f_{i'j'}
  \end{equation*} 
  \end{proof}